\documentclass[12pt]{amsart}
\calclayout
\usepackage[table]{xcolor}
\usepackage{times}
\usepackage{amssymb, amsmath}
\usepackage{amsthm}
\usepackage{tikz}
\usepackage{bm}
\usetikzlibrary{matrix,arrows}
\usepackage{enumitem}
\usepackage{booktabs}
\usepackage[section]{placeins}
\usepackage{subcaption}
\usepackage{multirow}
\usepackage{hyperref}
\usepackage{float}
\theoremstyle{plain}

\theoremstyle{remark}

\definecolor{Gray}{gray}{0.95}
\newcolumntype{g}{>{\columncolor{Gray}}c}

\graphicspath{{./images/}}

\begin{document}

\title{Synthetic Data for Feature Selection}

\author[F. Kamalov, H. Sulieman,  A. Cherukuri]{Firuz Kamalov$^1$$^{\boldsymbol{*}}$, Hana Sulieman$^2$, Aswani Cherukuri$^3$}

\address{$^{1}$ Department of Electrical Engineering\\
Canadian University Dubai, Dubai, UAE.}
\email{\textcolor[rgb]{0.00,0.00,0.84}{firuz@cud.ac.ae}}

\address{$^{2}$ Department of Mathematics and Statistics\\
 American University of Sharjah, Sharjah, UAE}
\email{\textcolor[rgb]{0.00,0.00,0.84}{hsulieman@aus.edu}}

\address{$^{3}$ School of IT and Engineering\\
 Vellore Institute of Technology, 
 Vellore, India}
\email{\textcolor[rgb]{0.00,0.00,0.84}{cherukuri@acm.org}}

%\author{Firuz Kamalov}
%\subjclass[2000]{6207}
%^\address{Mathematics Department, Canadian University of Dubai, Dubai, UAE}
%\email{firuz.kamalov@huskers.unl.edu}
%\date{\today}
\date{\today
\newline \indent $^{\boldsymbol{*}}$ Corresponding author}

\begin{abstract}
Feature selection is an important and active field of research in machine learning and data science. Our goal in this paper is to propose a collection of synthetic datasets that can be used as a common reference point for feature selection algorithms. Synthetic datasets allow for precise evaluation of selected features and control of the data parameters for comprehensive assessment. The proposed datasets are based on applications from electronics in order to mimic real life scenarios. To illustrate the utility of the proposed data we employ one of the datasets to test several popular feature selection algorithms. The datasets are made publicly available on GitHub and can be used by researchers to evaluate feature selection algorithms.
\end{abstract}
\maketitle

\section{Introduction}
Feature selection is a well investigated subject with a large amount of literature devoted to its study.  In an effort to improve on the existing results, researchers are continuously introducing new feature selection methods. The existence of such a large corpus of feature selection methods creates a challenge to compare and contrast different methods. Part of the challenge lies in the lack of agreement among researchers regarding benchmark datasets to be used for evaluating feature selection algorithms. Each team of researchers uses its own judgment and preference to decide on the data to be used for evaluating their proposed method. Given that feature selection methods are often tested on different datasets, it is hard to compare and analyze the reported results. 

In this paper, we attempt to standardize the evaluation process by introducing a collection of synthetic datasets that are designed specifically for the purpose of feature selection. Synthetic data offers several advantages over real-life data \cite{Jordon}. Unlike real-life data, the relevant features of synthetic data are known a priori. Thus, it is possible to directly evaluate a feature selection method with the exact knowledge of the correct features. In addition, synthetic data allows us to control the parameters of data and analyze the performance of feature selection methods under various conditions. By varying the number of irrelevant and correlated features we can investigate the precision of feature selection methods. Synthetic data also allows us to control the amount of random noise and study the corresponding effectiveness of the selection algorithms. Finally, by controlling the degrees of freedom of the target value function, we can measure the response of selection algorithms to nonlinear problems.

The proposed synthetic datasets are inspired by ideas from electrical engineering. In particular, we employ digital logic, electric circuits, and light-emitting diodes to design the synthetic datasets.
Digital logic is applied in  electronics and computer architecture as well as in robotics and other fields. Therefore, using digital logic allows us to construct realistic datasets for feature selection. Logic operations allow the construction of both linear and nonlinear target variables. In addition, redundant and correlated variables as well as noise can easily be implemented within synthetic data using digital logic. Similarly, electric circuits play a fundamental role in electrical engineering. Given the analog nature of electric circuits, we can generate continuous-valued data to simulate regression tasks.

While synthetic datasets have been used for feature selection in the past, their adoption as standard benchmark datasets has been limited \cite{Bolon, Chen, Kim}. This is partly due to the lack of datasets that are specifically designed for feature selection and not just general machine learning tasks. In addition, there does not exist a simple way for accessing these datasets. To overcome, these drawbacks the proposed synthetic data is purpose-built for feature selection. 
Furthermore, for convenience, the code for generating the proposed data is shared on GitHub. Thus, we hope to bring the evaluation process into a common framework to advance the research on feature selection algorithms.

Our paper is structured as follows. In Section 2, we provide a brief overview of the existing literature. In Section 3, we describe the details of the proposed synthetic datasets. In Section 4, we use one of the proposed datasets to evaluate the performance of feature selection methods. Section 5 concludes the paper.

\section{Literature}
Most of the existing synthetic data used in feature selection was originally designed for general machine learning tasks such as classification and regression. In particular, synthetic datasets based on Boolean features and operations have been utilized in the past, albeit in a limited capacity. The most well-known Boolean dataset is the XOR data which consists of two relevant features together with several irrelevant features \cite{Kim}. The target variable is determined by the formula $Y=X_1\oplus X_2$. Another commonly used Boolean data is based on the parity function. It consists of $n$ relevant features and the target variable is given by the rule $Y=1$ if the number of $X_i=1$ is odd and $Y=0$ otherwise \cite{Belanche}. In \cite{John}, the authors proposed the CorrAL dataset which consists of six Boolean features and the target variable is given by the formula $(x_1\land x_2) \lor (x_3\land x_4)$. Feature $X_5$ is irrelevant and feature $X_6$ is correlated with $Y$ by 75\%. 
An extension of the CorrAL dataset with 100 features was proposed in \cite{Kim}.
In \cite{Thrun}, the authors use six binary features to describe the task of a robot which is used as a basis for a synthetic dataset \cite{Bolon}. In \cite{Zhu}, the authors mimicked microarray data to create synthetic data with similar characteristics.

A synthetic dataset with continuous features and a binary target variable was proposed in \cite{Wang}, where the authors divided 100 features into 10 groups of highly correlated variables. Then the points where randomly assigned into two clusters using the Gaussian distribution. Another dataset with continuous features and a binary class value was given in \cite{Loscalzo}, where the target variable was decided as a linear function of equal weights of the relevant features. In \cite{Liu}, the authors used a similar approach but using a nonlinear neural network to generate the class labels.

Synthetic data has also been employed in nontraditional feature selection tasks such as unsupervised feature selection \cite{Panday} and dynamic feature selection \cite{Kaya}.
We note that not all synthetic data is well suited for feature selection \cite{Varol, Ward}. 

Perhaps the most well-known continuous-valued synthetic data are the Friedman datasets \cite{Breiman, Friedman}. The datasets consists of several relevant and irrelevant continuous-valued features. The target value is computed based on polynomial, rational, and trigonometric operations. The Friedman datasets were originally designed for testing regression models. Researchers have subsequently also used it in feature selection.
%---------------------------------------------------------------------------------------------------------------
%\section{Feature Relevance}
%Discuss feature relevance: weak relevance; strong relevance; fractional relevance.
%Feature relevance vs subset relevance: it is possible to have a subset of two correlated features that is more "informative" and/or accurate than a single relevant feature.
%Measures of relevance: mutual information, classifier accuracy.

%---------------------------------------------------------------------------------------------------------------
\section{Synthetic Data}
In this section, we describe the proposed synthetic data for feature selection. The datasets are presented in order from simple to complex. The summary of the proposed datasets is provided in Table \ref{dataset_details}. Every dataset consists of a total of 100 features and a target variable. The target value is completely determined by the relevant features. 

In our approach, we chose to include the same number of redundant features as the relevant ones. The redundant features are created either as a negation of the corresponding relevant feature in the case of binary features or as a linear function of the corresponding relevant feature in the case of continuous features. In addition, each dataset includes 2 correlated features. The correlated features are created by randomly  changing the value of the target variable in 30\% of the instances. Thus, the correlated features are correlated with the target variable 70\% of the time. Note that the correlated features have no causal relation with the target variable and are not desirable.
The remaining features are randomly generated to be irrelevant.

The proposed datasets cover a range of possibilities for the target variable including binary, multi-class, and continuous values. The number of observations is chosen to be small relative to the number of features. However, it can be adjusted as necessary. The details of each proposed dataset are presented in the subsequent sections.

\begin{table}[htb]
\centering
\begin{tabular}{lrrrrrr}
\toprule
Name & Relevant &  Redundant & Correlated & Irrelevant & Samples & Target\\
\midrule
ORAND & 3 &  3 & 2& 92&50 &binary \\
ANDOR & 4 &  4 &  2 & 90&50  &binary\\
ADDER& 3 &  3 &  2 & 92&50  &4-class\\
LED-16 & 16 &  16 & 2 & 66&180  &36-class\\
PRC &  5 &  5 & 2 & 88 & 500  &continuous\\
\bottomrule
\end{tabular}
\caption{Summary of the proposed datasets.}
\label{dataset_details}
\end{table}

The irrelevant variables are created using a random number generator. To maintain consistency a fixed random seed is used in generating the irrelevant variables. All the datasets were created using Python programming language. The datasets and the corresponding code are publicly available on  \href{https://github.com/group-automorphism/synthetic_data}{GitHub}. 

%---------------------------------------------------------------------------------------------------------------
\subsection{ORAND}
\begin{figure}[H]
\center
\includegraphics[trim={4cm 11cm 14cm 6cm},clip, width=0.5\textwidth]{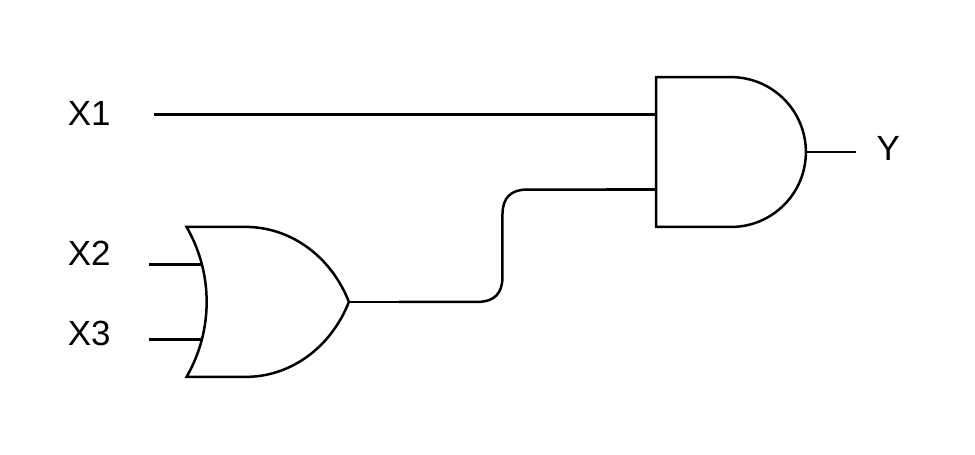}
\caption{The ORAND circuit diagram.}
\label{orand}
\end{figure}

The proposed dataset is based on a two-layer circuit as shown in Figure \ref{orand}. The first layer consists of an OR gate, while the second layer consists of an AND gate.

The ORAND dataset contains three relevant features $X_1, X_2$, and $X_3$. The target variable $Y$ is calculated via the following formula:
\begin{equation}
Y=X_1\land (X_2\lor X_3).
\end{equation}
In addition to the three relevant variables, we add three redundant (correlated) variables - one for each of $X_1, X_2$, and $X_3$. We also add 2 features that randomly match the target variable in 70\% of the instances. In the end, we include
$N_I=92$ irrelevant features which are randomly generated according to the Bernoulli process. We obtain a synthetic dataset consisting of 3 relevant, 3 redundant, 2 correlated, and  92 irrelevant features. There are a total $2^{100}$ possible feature value combinations of which we select $n=50$ samples. As it is often the case with high dimensional data, the number of features is high relative to the number of observations. While the default number of samples is 50, it can be changed as necessary. In fact, both the number of observations  and the number of irrelevant features can be varied to analyze the performance of a feature selection algorithm under different conditions. However, it is strongly advised to employ the proposed datasets under the default settings, at least once, to obtain a benchmark reference result.

%---------------------------------------------------------------------------------------------------------------
\subsection{ANDOR} \label{sec_andor}

\begin{figure}[H]
\center
\includegraphics[trim={6cm 9cm 14cm 6cm},clip, width=0.4\textwidth]{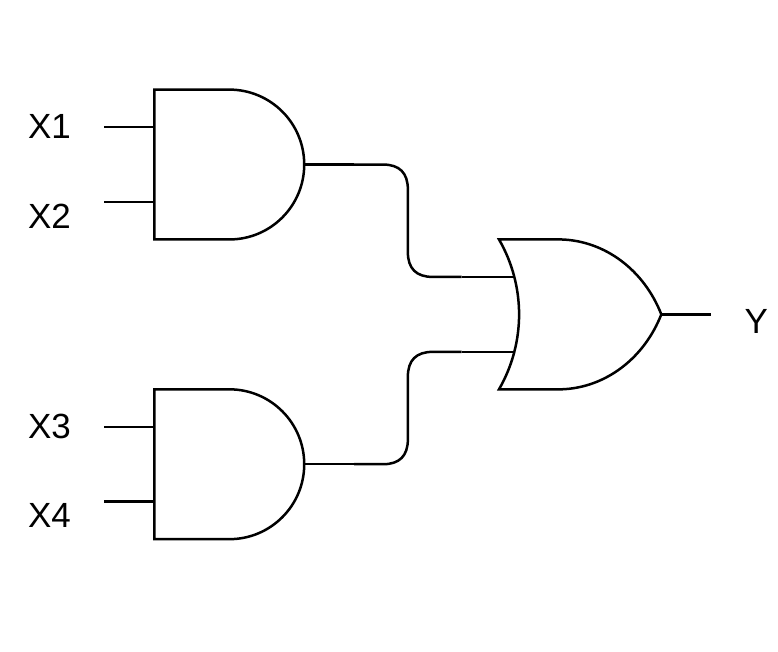}
\caption{The ANDOR circuit diagram.}
\label{and2or}
\end{figure}

The proposed dataset is based on a two-layer circuit as shown in Figure \ref{and2or}. The first layer consists of two AND gates, while the second layer consists of a single OR gate.

The ANDOR dataset contains four relevant features $X_1, X_2, X_3$, and $X_4$. The target variable $Y$ is calculated via the following formula:
\begin{equation}\label{eq:andor}
Y=(X_1\land X_2)\lor (X_3\land X_4).
\end{equation}
The expression in Equation \ref{eq:andor} can be viewed as the sum of products. 
In addition to the four relevant variables, we add four redundant (correlated) variables - one for each of $X_1, X_2, X_3$, and $X_4$. 
We also add 2 features that randomly match with the target variable 70\% of the time. Finally, we include
 $N_I=90$ irrelevant features which are randomly generated according to the Bernoulli process. We use the features to randomly generate $n=50$ samples.
We obtain a synthetic dataset consisting of 4 relevant, 4 redundant, 2 correlated, and 90 irrelevant features and 50 samples.
As with the ORAND dataset, the number of irrelevant features $N_I$ and the number of samples $n$ can be changed to evaluate a feature selection algorithm under different scenarios.

The ANDOR dataset  can be expanded by increasing the number of relevant variables up to $m$ in the following manner:
\begin{equation}
Y=(X_1\land X_2)\lor (X_3\land X_4)\dots \lor (X_{n-1}\land X_m).
\end{equation}
Additional redundant and irrelevant features can be included in similar fashion as above.

%---------------------------------------------------------------------------------------------------------------
\subsection{ADDER}
The ADDER dataset is based on the eponymous adder circuit. It is a multi-class target dataset. The full adder takes three inputs $X_1, X_2$, and $X_3$ and produces two outputs $Y_1$ and $Y_2$. The outputs are calculated according to the following formulae:

\begin{equation}
\begin{split}
Y_1 &= X_1\oplus X_2\oplus X_3\\
Y_2 &= (X_1\land X_2)\lor (X_3\land(X_1\oplus X_2))
\end{split}
\end{equation}
By combining the values of  $Y_1$ and $Y_2$ into a single target variable $Y = (Y_1, Y_2)$, we obtain a 4-class target variable: $Y=\{(0,0), (0,1), (1,0), (1,1)\}$. As usual, we add redundant  features - one for each of $X_1, X_2$, and $X_3$ - 2 correlated features, and $N_I=92$ irrelevant features. We randomly generate $n=50$ samples based on the full set of features. The final dataset consists of 3 relevant, 3 redundant, 2 correlated, and 92 irrelevant features and 50 observations. As mentioned above, unlike ORAND and ANDOR, ADDER is a multi-class dataset with a 4-class target variable.

%---------------------------------------------------------------------------------------------------------------
\subsection{LED-16}
\begin{figure}[htb]
\center
\includegraphics[trim={0cm 0cm 0cm 0cm},clip, width=0.12\textwidth]{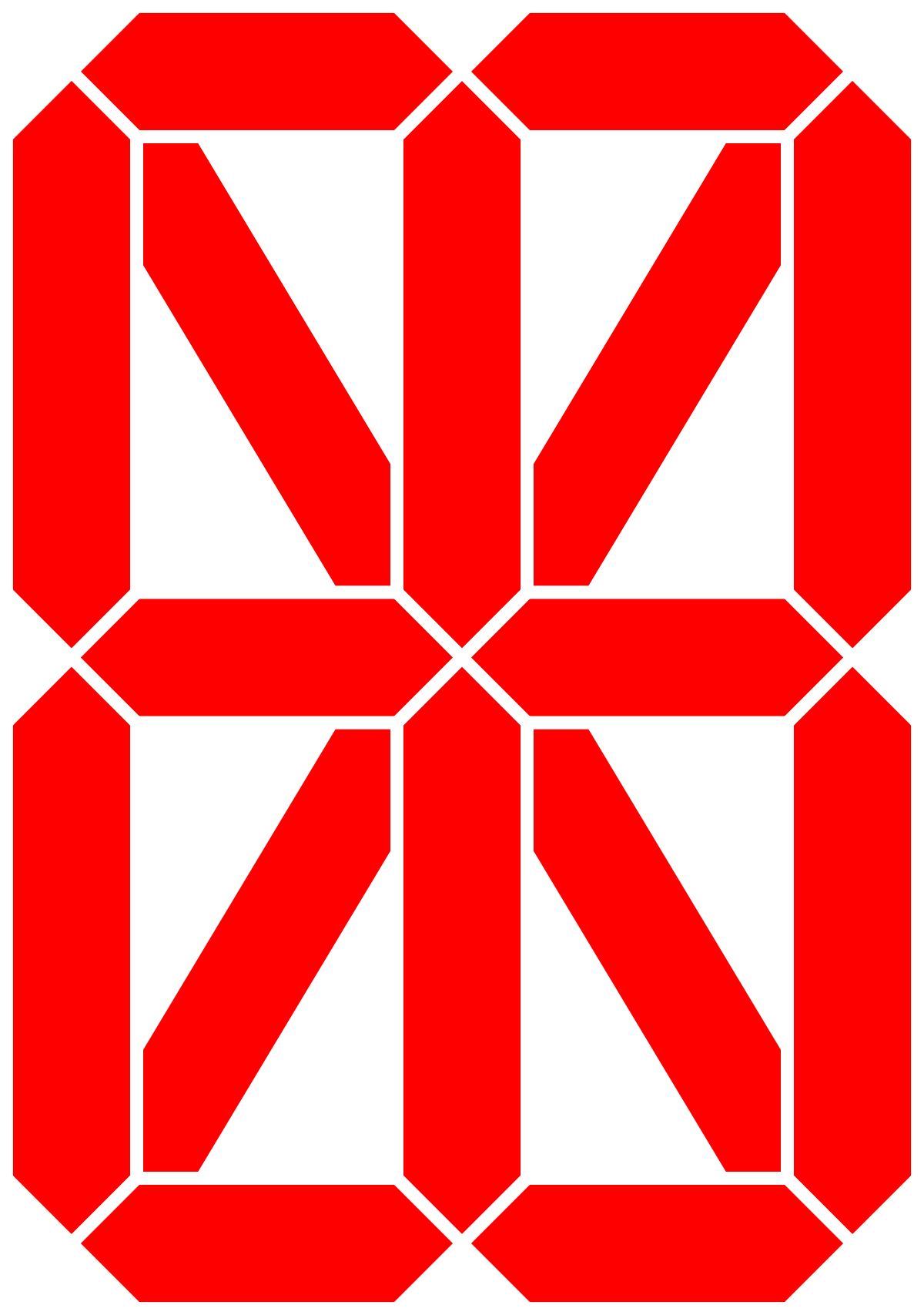}
\caption{The 16-segment LED display.}
\label{16-seg}
\end{figure}

The LED-16 dataset is based on the 16-segment display configuration shown in Figure \ref{16-seg}. The 16-segment configuration allows to display all 26 letters of the English alphabet as well as all the digits 0-9. Each segment represents a binary feature: on/off. The target variable is the alpha-numeric value displayed by the segments. Thus, the target variable can take 36 different values. We add 16 redundant features - one for each relevant feature (segment). We also include 2 correlated and 66 irrelevant features with randomly generated binary values. 

The dataset contains 180 observations - 5 samples for each target value. In particular, for each target value, the values of the relevant and redundant features are determined by the display configuration, while the values of remaining features are randomly generated. In summary, the LED-16 dataset consists of 16 relevant, 16 redundant, 2 correlated, and 66 irrelevant features and 180 observations. It is a multi-class dataset with a 36-class target variable.

\begin{figure}[htb]
\center
\includegraphics[trim={0cm 0cm 0cm 0cm},clip, width=1\textwidth]{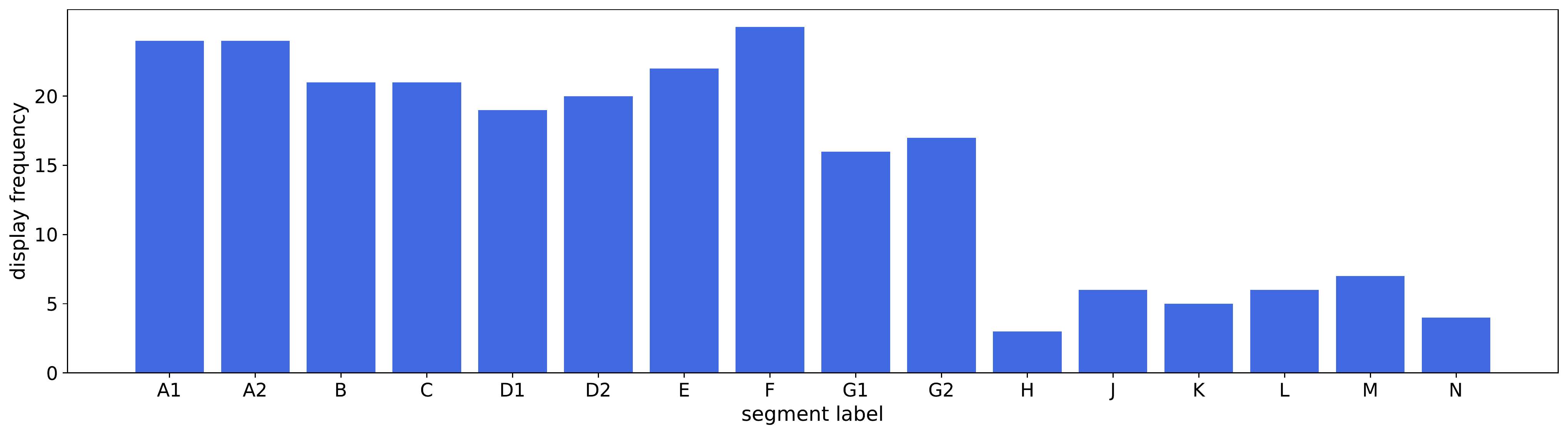}
\caption{Frequency distribution of the LED-16 segments over 36 alphanumeric characters.}
\label{segments}
\end{figure}

As shown in Figure \ref{segments}, some LED segments are used more frequently than others. For example, the segment F is utilized in the majority of configurations, while the segment H is used in only a few times. While both segments are relevant, they have different levels of significance. The LED-16 dataset allows to evaluate feature selection algorithms on the basis of their sensitivity to different level of feature relevance. A robust selection algorithm should identify both strongly and weakly relevant features.

%---------------------------------------------------------------------------------------------------------------
\subsection{PRC}
The proposed dataset is based on the parallel resistor circuit. Given a set of resistors $\{R_1, R_2, ..., R_N\}$ that are connected in parallel, the total resistance $R_T$ is given by the following equation:
\begin{equation}\label{prc}
\frac{1}{R_T} = \sum_{i=1}^N \frac{1}{R_i}.
\end{equation}

For the proposed PRC dataset, we use 5 parallel connected resistors (relevant features). The relevant features $\{X_i\}_{i=1}^5$ are generated according to the Gaussian distribution with mean $\mu_i=10$ and standard deviation $\sigma_i=i, i=1, 2, ...,5$. The target variable $Y=R_T$ is calculated according to Equation \ref{prc}. Since the parallel resistors as well as the total resistance are continuous valued variables, the PRC dataset implies a regression task. 

We generate 5 redundant, 2 correlated, and 88 irrelevant features to include in the dataset. The redundant variables are created as linear transformations of the relevant variables. The correlated variables are created by adding a small amount of noise perturbation to the target variable. Half of the irrelevant variables are generated according to the same Gaussian distribution as the relevant variables. In particular, for each $i=1, 2, ...,5$, we generated nine random features according to the Gaussian distribution $\mathcal{N}(\mu_i=10, \sigma_i=i)$.
The remaining half of the irrelevant features are randomly generated according to the uniform distribution $\mathcal{U}(0, 1)$. 

Since the target variable $Y$ is continuous valued, we generate more observations for the PCR dataset than for the classification data described in the previous sections. In particular, we generate 500 samples using the full feature set, where the target variable is calculated via  Equation \ref{prc} based on the features $\{X_i\}_{i=1}^5$.

%---------------------------------------------------------------------------------------------------------------
\subsection{Comparator}
A comparator circuit compares two inputs - usually voltages or currents - and outputs either 1 or 0 depending on which input is greater. It is often used to check if a single input is above a certain threshold. In this manner, a comparator can employed to convert an analog (continuous) signal into a digital (binary) output.

Since a comparator converts a continuous signal into binary, it can be employed in conjunction with the digital circuits described in previous sections including ORAND, ANDOR, and ADDER. For instance, one or more input variables in the ORAND circuit can be continuous-valued by using a comparator as a preprocessing gate. Thus, datasets with binary features can be extended to include continuous features, which increases the number of available synthetic datasets for feature selection.

%---------------------------------------------------------------------------------------------------------------
\section{Feature selection based on synthetic data}
To illustrate the application of the proposed synthetic data we evaluate the performance of standard feature selection algorithms on the ANDOR dataset. Since the relevant features in synthetic data are known, we can directly analyze the effectiveness of the algorithms. Synthetic data also gives an option to control the number of samples and the level of noise in the data which enables researchers better understand the performance of the algorithms under different scenarios.

\subsection{Methodology}
We use the ANDOR dataset described in Section \ref{sec_andor} to evaluate the feature selection algorithms. The dataset consists of 4 relevant, 4 redundant, 2 correlated, and 90 irrelevant features and a binary target class. The redundant variables are the negations of the relevant variables. The correlated variables are correlated with the target variable at 70\%. 
To observe the effect of the sample size we consider the datasets with 50 and 20 samples. 

To account for random variations, we generate the ANDOR dataset using 10 different random seeds. The feature selection algorithms are applied to each of the 10 datasets. The results are averaged and presented for analysis.

\subsection{Models}
We consider univariate algorithms, recursive feature elimination (RFE), and model-based algorithms for feature selection. In the univariate approach, we employ $\chi^2$ to measure the  relationship between a feature and the target variable. In the RFE approach, we use the support vector classifier (SVC) with linear kernel. The SVC model is iteratively fit to the data and at each stage the variable with the lowest coefficient is discarded. In the model-based approach, we use SVC with the $L_1$-penalty (lasso) and the extra-trees classifier. The lasso model automatically discards the irrelevant features as part of the fitting process. The extra-trees classifier is used to calculate the reduction in impurity of each feature during the fitting process.

\subsection{Results}
Four feature selection algorithms - univariate ($\chi^2$), RFE, lasso, and extreme trees - are tested based on the ANDOR dataset.
The results show that the feature selection methods perform relatively well in distinguishing between the relevant and the irrelevant features. However, the selection algorithms struggle to separate the relevant features from the redundant and correlated features. Furthermore, a decrease in the sample size leads to a lower distinction between the relevant and irrelevant features.

\begin{figure}[ht]
\begin{subfigure}{1\textwidth}
  \centering
  % include first image
  \includegraphics[width=1\linewidth]{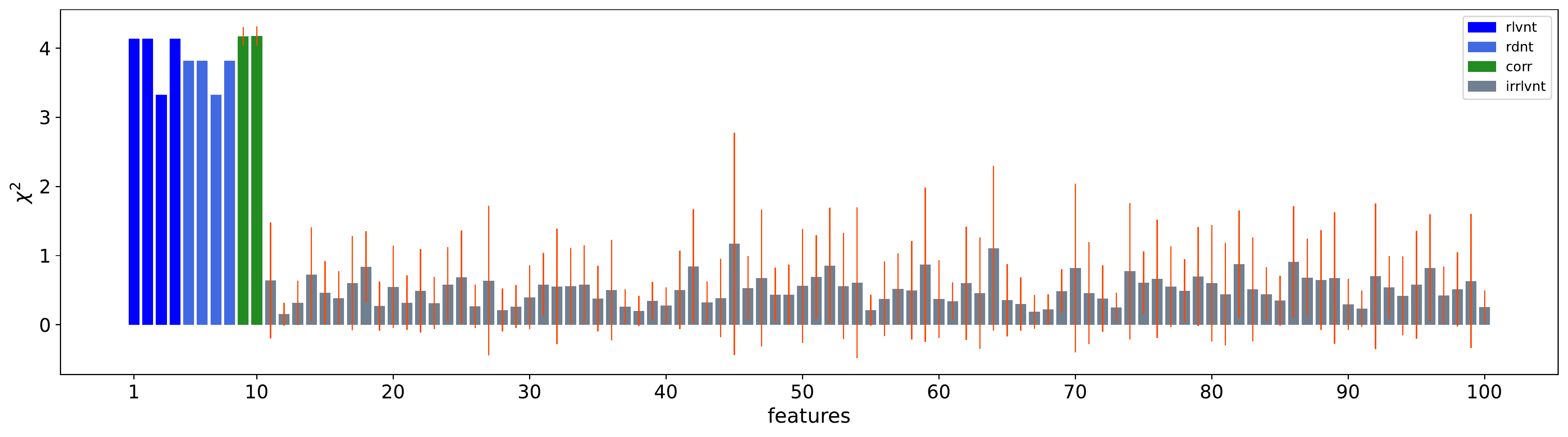}  
  \caption{The ANDOR dataset with 50 samples.}
  \label{uni_50}
\end{subfigure}
\\
\begin{subfigure}{1\textwidth}
  \centering
  % include second image
  \includegraphics[width=1\linewidth]{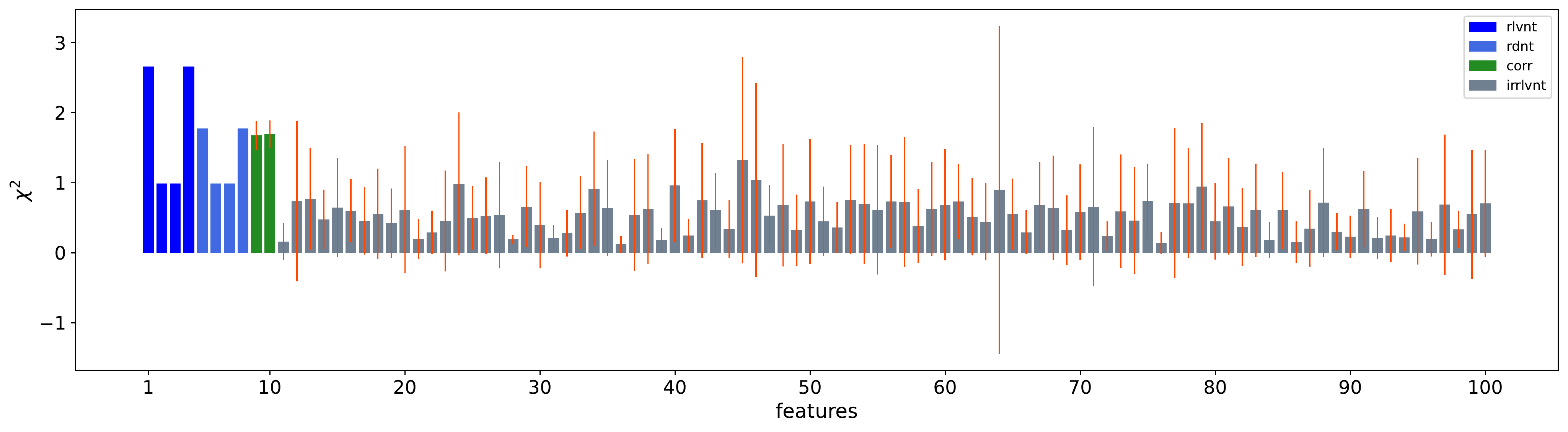}  
  \caption{The ANDOR dataset with 20 samples.}
  \label{uni_20}
\end{subfigure}
\caption{The mean $\chi^2$-scores between the features and the target variable. The red line segments show the standard deviations of the scores.}
\label{uni_fig}
\end{figure}

The results of the univariate approach for feature selection are presented in Figure \ref{uni_fig}. The average $\chi^2$-scores between the features and the target variable over 10 runs are provided. Since the correlated and irrelevant features are randomly generated, there exists variation in the results between the runs. To show the variations between the runs the standard deviation of $\chi^2$-scores are represented by the red line segments.

As shown in Figure \ref{uni_50}, in case when the sample size is $n=50$, the algorithm assigns significantly higher $\chi^2$-scores to the relevant features than the irrelevant features. On the other hand, it also assigns high scores to the redundant and correlated features. Thus, while the algorithm is able to distinguish the relevant features from the irrelevant features, it fails to distinguish the relevant features from the redundant and correlated features. As shown in Figure \ref{uni_20}, in case when the sample size is $n=20$, the difference in scores between the relevant and irrelevant features is considerably lower than in the case $n=50$. Given the large variation of the irrelevant features, it is highly likely that at least some of the irrelevant features may obtain a higher $\chi^2$-score than the relevant features. The features $X_2$ and $X_3$ are particularly susceptible to being overlooked.

The results of the RFE approach are presented in Table \ref{rfe_tab}. The tables present RFE feature rankings for the relevant ($X_1$ to $X_4$) and redundant ($X_5$ to $X_8$) features. The left and right tables correspond to the data of size $n=50$ and $n=20$, respectively. The rows in the tables correspond to the datasets generated using different random seeds.  The last row $\boldsymbol{\tilde{r}}$ provides the median rankings of the features across 10 runs. The results show that the RFE approach is less effective than the univariate approach. There is a significant variation in the rankings between different versions of the dataset which suggests that RFE is not a stable method. To summarize the results we consider the median rankings.
In case $n=50$, 6 features achieved median ranking in the top 10, but only one feature $X_8$ is in top 4. Thus, if we were to select the top 4 features - since that is the number of relevant variables - only one feature would be correctly selected.
In case the $n=20$, the algorithm performs poorly with only one feature in top 10 and none in the top 4. It is unsurprising that the algorithm is less effective when using a small sample size.

\begin{table}[htb]
\centering
\subfloat[The ANDOR dataset with 50 samples.]{
\begin{tabular}{lrrrrrrrr}
\toprule
 \textit{v}& $X_1$ & $X_2$ & $X_3$ & $X_4$ & $X_5$ & $X_6$ & $X_7$ & $X_8$ \\
 \midrule
1 & 15 & 14 & 4 & 13 & 3 & 2 & 18 & 1 \\
2 & 12 & 5 & 23 & 10 & 9 & 8 & 11 & 3 \\
3 & 12 & 16 & 5 & 6 & 3 & 1 & 2 & 9 \\
4 & 22 & 13 & 5 & 3 & 6 & 17 & 11 & 14 \\
5 & 2 & 7 & 9 & 6 & 8 & 4 & 3 & 5 \\
6 & 13 & 4 & 7 & 6 & 5 & 14 & 2 & 1 \\
7 & 14 & 5 & 4 & 15 & 3 & 11 & 12 & 2 \\
8 & 11 & 12 & 9 & 6 & 5 & 2 & 4 & 3 \\
9 & 6 & 13 & 12 & 4 & 14 & 7 & 10 & 8 \\
10 & 7 & 11 & 13 & 14 & 3 & 8 & 10 & 6 \\
\midrule
$\boldsymbol{\tilde{r}}$ & \textbf{12} & \textbf{12} &  \textbf{8} &  \textbf{6} & \textbf{5} & \textbf{8} & \textbf{10} &  \textbf{4}\\
\bottomrule
\end{tabular}}
\quad
\subfloat[The ANDOR dataset with 20 samples.]{
\begin{tabular}{lrrrrrrr}
\toprule
$X_1$ & $X_2$ & $X_3$ & $X_4$ & $X_5$ & $X_6$ & $X_7$ & $X_8$ \\
 \midrule
5 & 8 & 20 & 6 & 34 & 13 & 39 & 7 \\
29 & 17 & 8 & 5 & 13 & 9 & 16 & 14 \\
12 & 7 & 50 & 17 & 4 & 22 & 53 & 8 \\
5 & 33 & 39 & 12 & 14 & 29 & 46 & 3 \\
16 & 35 & 32 & 9 & 22 & 13 & 26 & 11 \\
36 & 11 & 17 & 5 & 28 & 4 & 21 & 6 \\
18 & 22 & 37 & 11 & 10 & 35 & 33 & 5 \\
14 & 6 & 17 & 12 & 18 & 28 & 4 & 10 \\
13 & 43 & 7 & 42 & 9 & 37 & 23 & 30 \\
23 & 19 & 27 & 11 & 6 & 20 & 31 & 3 \\
\midrule
\textbf{15} & \textbf{18} & \textbf{24} & \textbf{11} & \textbf{14} & \textbf{21} & \textbf{29} &  \textbf{8}\\
\bottomrule
\end{tabular}}
\caption{The rankings of the relevant ($X_1$ to $X_4$) and redundant ($X_5$ to $X_8$) features for each of the 10 versions of the ANDOR dataset using the RFE algorithm.}
\label{rfe_tab}
\end{table}

\begin{figure}[htb]
\begin{subfigure}{1\textwidth}
  \centering
  % include first image
  \includegraphics[width=1\linewidth]{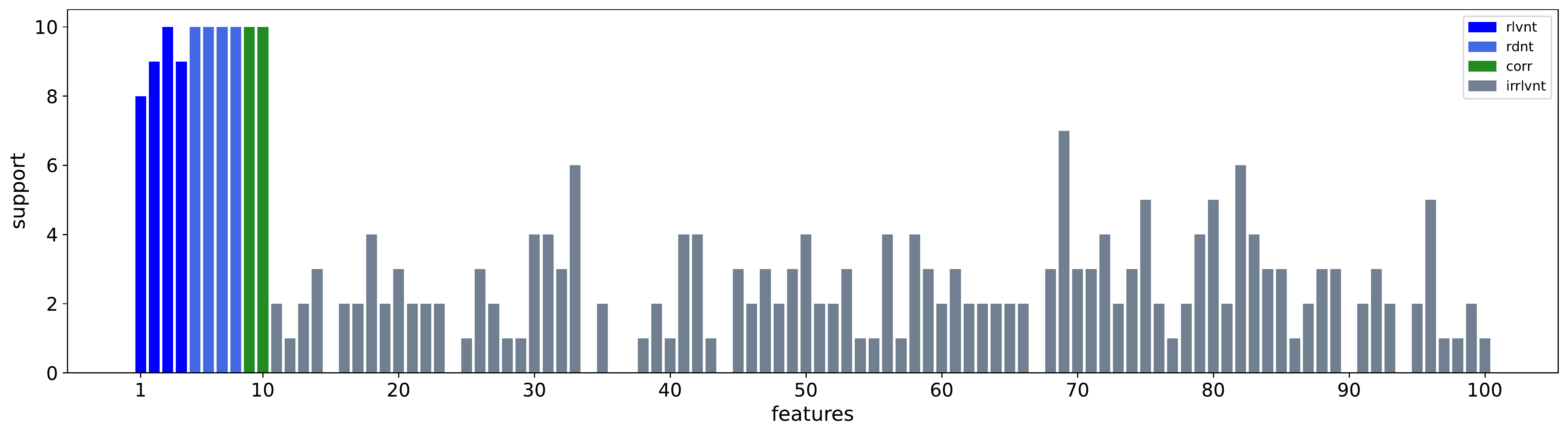}  
  \caption{The ANDOR dataset with 50 samples.}
  \label{lsvc_50}
\end{subfigure}
\\
\begin{subfigure}{1\textwidth}
  \centering
  % include second image
  \includegraphics[width=1\linewidth]{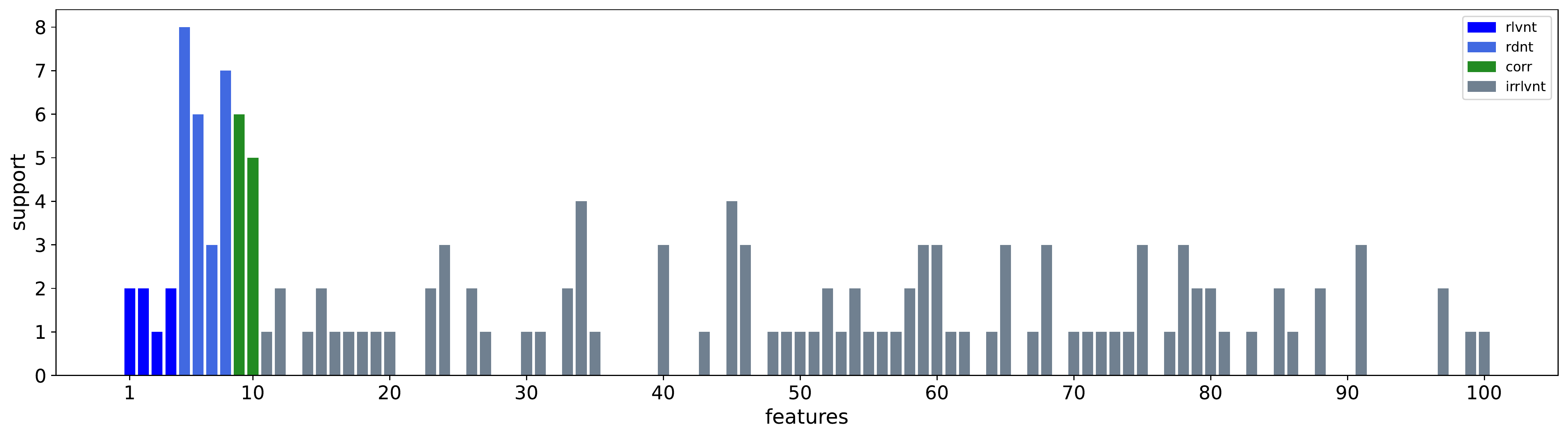}  
  \caption{The ANDOR dataset with 20 samples.}
  \label{lsvc_20}
\end{subfigure}
\caption{The number of times each feature was chosen in the lasso model for 10 versions of the dataset.}
\label{lsvc_fig}
\end{figure}

The results of the lasso-based approach  in case the $n=50$ are presented in Figure \ref{lsvc_50}. The $y$-axis represents the number of times a feature was selected in the lasso model. As shown in the figure, the lasso approach performed well in this case. The relevant features achieve significantly higher scores than the irrelevant features. However, the redundant and correlated features also achieve high scores. Thus, while the lasso approach is effective at distinguishing between the relevant and irrelevant features, it fails to differentiate the relevant features from the redundant and correlated features. The results in the case $n=20$ are expectedly less robust. As shown in Figure \ref{lsvc_20}, the scores for the relevant features are significantly lower than those for the correlated features. In addition, several irrelevant features have higher scores than the relevant features. On the other hand, three of the redundant features have high scores which can be an acceptable substitute for the relevant features.

\begin{figure}[htb]
\begin{subfigure}{1\textwidth}
  \centering
  % include first image
  \includegraphics[width=1\linewidth]{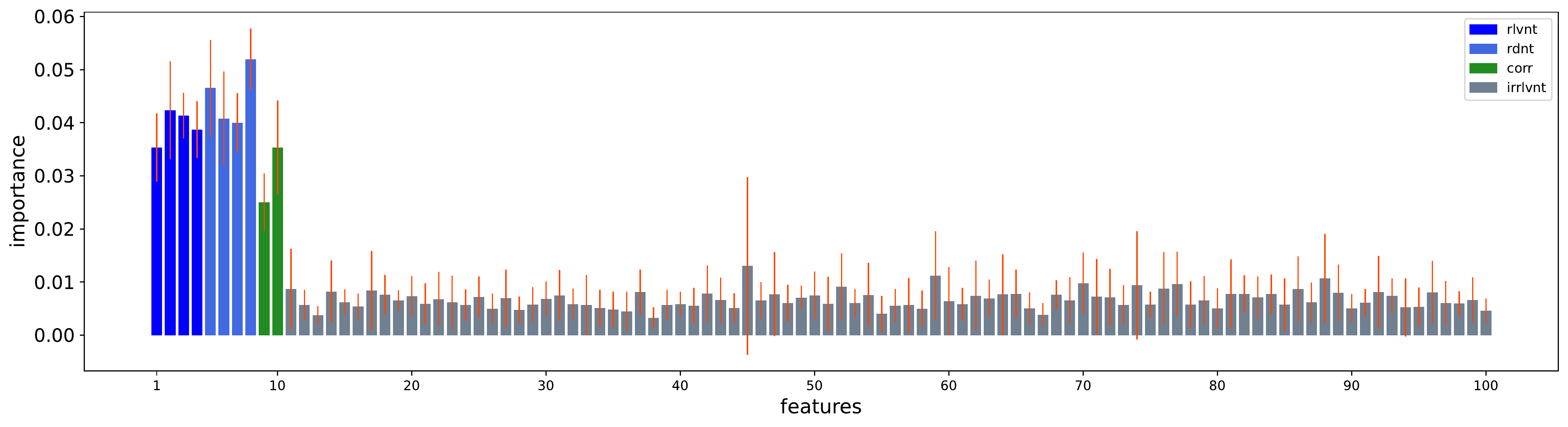}  
  \caption{The ANDOR dataset with 50 samples.}
  \label{tree_50}
\end{subfigure}
\\
\begin{subfigure}{1\textwidth}
  \centering
  % include second image
  \includegraphics[width=1\linewidth]{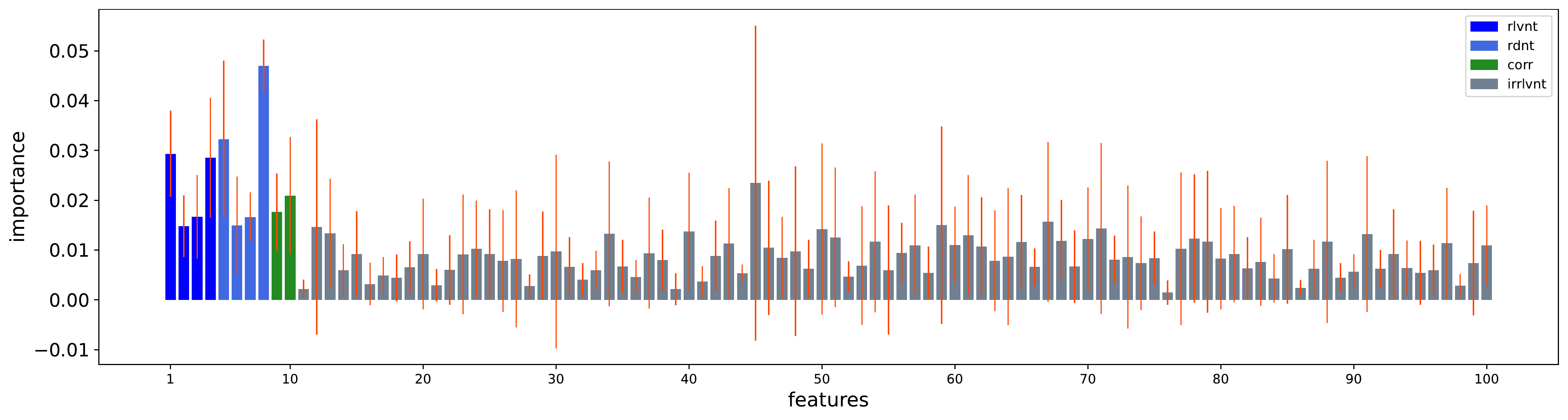}  
  \caption{The ANDOR dataset with 20 samples.}
  \label{tree_20}
\end{subfigure}
\caption{The average decrease in impurity using the extra-tree classifier.}
\label{tree_fig}
\end{figure}

The results of the approach based on extra-trees are presented in Figure \ref{tree_fig}. The $y$-axis indicates the average decrease in impurity corresponding to a feature. As shown in Figure \ref{tree_50}, in the case of $n=50$, the algorithm performs better than the previous approaches by assigning higher level of importance to relevant features than the irrelevant and correlated features. We also note that the irrelevant features have low variance. So they are less likely to be selected by chance. However, the algorithm does not distinguish well between the relevant and redundant variables. As expected, the algorithm performs worse in the case $n=20$. While the importance of the relevant variables $X_1$ and $X_4$ remains significantly above the irrelevant variables, the gap between the relevant variables $X_2$ and $X_3$ and the rest of the features decreased considerably. In addition, the variance of the irrelevant variables increased which increases the likelihood of an irrelevant variable achieving a high level importance purely by chance.

%--------------------------------------------------------------------------------------------------------
\section{Conclusion}

Our aim in this paper was to promote common benchmarking of feature selection algorithms. To this end, we proposed a set of synthetic data that can be used as a universal reference point.
The proposed synthetic datasets are motivated by ideas from electrical engineering. By mimicking real-life scenarios our goal was to produce data that can provide sensible evaluation of feature selection algorithms. Synthetic data has several advantages including the knowledge of the relevant features and the ability to vary the parameters of the data. As a result, synthetic data allows for a comprehensive evaluation of feature selection algorithms. 

To promote the adoption of common benchmarking, the proposed datasets are made publicly available on \href{https://github.com/group-automorphism/synthetic_data}{GitHub}. We encourage researchers to test their algorithms on these datasets. For convenience, we provide the code for generating the data which can be used to modify the parameters of the data as needed. However, we strongly advise to use the data under the default settings at least once.

In the future, more synthetic datasets inspired by different sources can be added to the proposed collection. In particular, synthetic data mimicking the gene data in medical research would be desirable. Data for regression tasks with continuous features would also enrich the existing collection.

%--------------------------------------------------------------------------------------------------------

\end{document}